\documentclass[10pt,journal,compsoc]{IEEEtran}

\usepackage{amsmath}
\ifCLASSOPTIONcompsoc
  \usepackage[nocompress]{cite}
\else
  \usepackage{cite}
\fi

\ifCLASSINFOpdf
  \usepackage[pdftex]{graphicx}
\else
  \usepackage[dvips]{graphicx}
\fi

\usepackage{url}

\usepackage{algorithmic}
\usepackage{algorithm}
\usepackage{array}
\usepackage[caption=false,font=normalsize,labelfont=sf,textfont=sf]{subfig}
\usepackage{stfloats}
\usepackage{booktabs}
\usepackage{xcolor}
\usepackage{amssymb}
\usepackage{pifont}
\usepackage{xspace}
\newcommand{\datasetname}{\textcolor{black}{RICH-CAT}\xspace}
\newcommand{\modelname}{\textcolor{black}{CATMO}\xspace}

\usepackage{multirow}
\usepackage{multicol}
\definecolor{Strawberry}{rgb}{1,0.26,0.64}

\newcommand{\et}[2]{${#1}^{\pm{#2}}$}

\newcommand{\etr}[2]{$\textcolor{red}{{#1}}^{\pm{#2}}$}
\newcommand{\etbb}[2]{$\textcolor{blue}{{#1}}^{\pm{#2}}$}

\usepackage[colorlinks,linkcolor=blue]{hyperref}
\usepackage{ragged2e}

\begin{document}
%
\title{Contact-aware Human Motion Generation from Textual Descriptions}

\author{Sihan~Ma,
        Qiong~Cao, 
        Jing~Zhang~\IEEEmembership{Senior Member,~IEEE,}
        and~Dacheng~Tao,~\IEEEmembership{Fellow,~IEEE}
\thanks{Sihan Ma and Jing Zhang are affiliated with The University of Sydney, Australia. Qiong Cao is associated with JD Explore Academy, China. Dacheng Tao is affiliated with Nanyang Technological University, Singapore.\protect\\
E-mail: sima7436@uni.sydney.edu.au, \{mathqiong2012, jingzhang.cv, dacheng.tao\}@gmail.com\protect\\
Corresponding authors: Dacheng Tao and Jing Zhang.
}
}

\markboth{Journal of \LaTeX\ Class Files,~Vol.~14, No.~8, August~2015}%
{Shell \MakeLowercase{\textit{et al.}}: Bare Demo of IEEEtran.cls for Computer Society Journals}

\IEEEtitleabstractindextext{%
\begin{abstract}
\justifying
This paper addresses the problem of generating 3D interactive human motion from text. 
Given a textual description depicting the actions of different body parts in contact with static objects, we synthesize sequences of 3D body poses that are visually natural and physically plausible.
Yet, this task poses a significant challenge due to the inadequate consideration of interactions by physical contacts in both motion and textual descriptions, leading to unnatural and implausible sequences. 
To tackle this challenge, we create a novel dataset named \datasetname, representing ``\textbf{C}ontact-\textbf{A}ware \textbf{T}exts'' constructed from the RICH dataset~\cite{rich}. \datasetname comprises high-quality motion, accurate human-object contact labels, and detailed textual descriptions, encompassing over 8,500 motion-text pairs across 26 indoor/outdoor actions. 
Leveraging \datasetname, we propose a novel approach named \modelname for text-driven interactive human motion synthesis that explicitly integrates human body contacts as evidence. We employ two VQ-VAE models to encode motion and body contact sequences into distinct yet complementary latent spaces and an intertwined GPT for generating human motions and contacts in a mutually conditioned manner. Additionally, we introduce a pre-trained text encoder to learn textual embeddings that better discriminate among various contact types, allowing for more precise control over synthesized motions and contacts. 
Our experiments demonstrate the superior performance of our approach compared to existing text-to-motion methods, producing stable, contact-aware motion sequences. Code and data will be available for research purposes at \url{https://xymsh.github.io/RICH-CAT}.
\end{abstract}

}

\maketitle

\IEEEdisplaynontitleabstractindextext

\IEEEpeerreviewmaketitle

\begin{figure*}
    \centering
    \includegraphics[width=\linewidth]{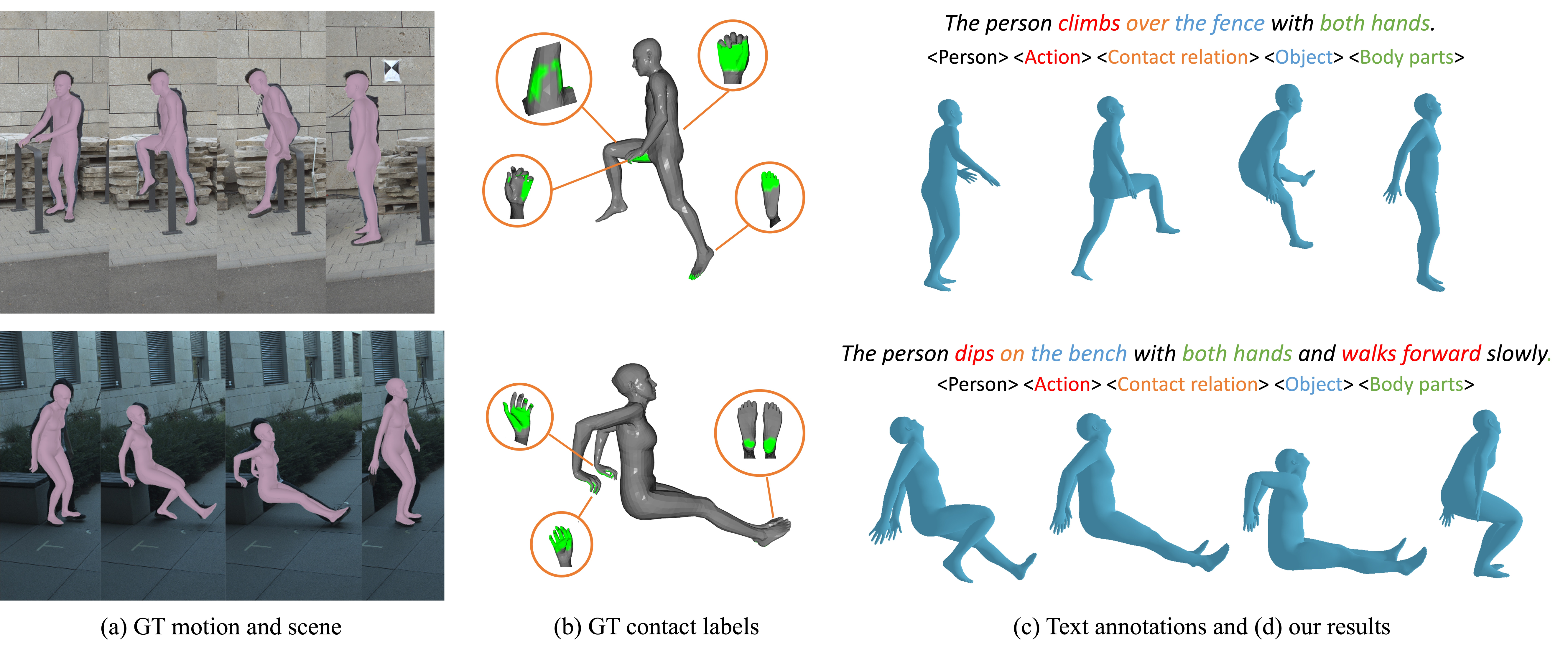}
    \caption{We address the problem of text-driven 3D interactive human motion generation from both data and algorithmic perspectives. \datasetname is a novel dataset featuring (a) high-quality motion, (b) accurate contact labels, and (c) interactive textual descriptions that specify different body parts interacting with various static objects. Using it, we introduce a novel approach named \modelname to learn the complex human motion dynamics that incorporate interaction semantics provided by contact, producing natural and plausible 3D human motions of action (d).}
    \label{fig:teaser}
\end{figure*}


\IEEEraisesectionheading{\section{Introduction}\label{sec:introduction}}

\IEEEPARstart{S}{ynthesizing} 3D human motion that makes contact with its surrounding environment is pivotal for simulating daily behaviors that individuals perform in the real world such as exercising or cooking. This is particularly critical in a variety of applications including humanoid robots, AR/VR and action games. The creation of interactive human behaviors from textual descriptions not only offers significant convenience but also extends the boundaries of these applications. 

This paper addresses the problem of text-driven 3D interactive human motion generation. Given a textual description illustrating the actions of various body parts in contact with static objects, we synthesize sequences of visually natural and physically plausible 3D body poses. The problem of text-driven motion generation~\cite{humanml3d,language2pose,motionclip,motiondiffuse,t2mgpt,mdm,mld} has seen considerable research interest.
However, existing methods exploring this domain lack comprehensive modeling of human motion with contacts in the real world, e.g., synthesizing only hand-grasping actions that make contact with small dynamic objects~\cite{imos}, limiting to a few types of predefined contact such as sitting on a chair~\cite{t2mgpt, mdm, mld} or overlooking the vertex-level contact labels by only considering collision~\cite{humanise}. In contrast, our focus is on a significantly more complex task, which requires accurate capturing of human movement in alignment with a textual description detailing the action types, interacting 3D body parts and objects.

To achieve this, we must learn a cross-modal mapping between text and interactive body motion, which is very challenging in practice for several reasons. First, there is a dearth of datasets that provide rich contextual 3D human motion interacting with various objects, accompanied by textual descriptions and accurate vertex-level contact labels. Second, considering the intricacy of body parts physically interacting with objects, depicting 3D interactive human actions based on text descriptions necessitates more than simple posture descriptions such as ``leans on the fence''. A more detailed illustration like ``Leaning on the fence with the left hand'' conveys the details. Lastly, as different body parts correspond to varying degrees with textual descriptions, it is difficult to model the cross-modal mapping and generate natural interactive body motions without explicit contact modeling.

We address the above challenges from both data and algorithmic perspectives and learn to model the complex interaction dynamics in a data-driven way. First, to overcome the issue of data scarcity and limited posture descriptions, we expand upon the RICH dataset~\cite{rich} by building a new set called \datasetname, featuring over 8,500 complex human-object motion and contact sequences belonging to 26 different in/outdoor action types. Each sequence is paired with a detailed description, generated by our automatic textual description generation technique, detailing different body parts engaged in interacting with objects. This provides a more precise portrayal of interactive human actions. 

Then, we investigate the translation from text to 3D interactive human motion. We propose \modelname, a novel approach that generates realistic 3D human motion in close contact with objects from a detailed textual description. Specifically, we utilize two independent VQ-VAEs to encode the motion and contact modalities into distinct yet complementary latent spaces, thereby capturing more intricate representations of each modality and providing each aspect with full expression. Next, to predict the 3D interactive human motion, we propose an intertwined GPT architecture that produces motion and contact in a twisted auto-regressive way, explicitly incorporating contact information into the motion generation process. With the learned Motion VQ-VAE decoder, the output from intertwined GPT is decoded, yielding a sequence of 3D poses with physically plausible contacts. Furthermore, we present a pre-trained text-motion model that effectively decomposes interactive textual descriptions and aligns them with the generated motion sequence by alignment loss. 

To the best of our knowledge, our proposed method is the first approach to explicitly address the problem of generating realistic 3D interactive human motion from textual descriptions depicting detailed interactions by contacts. We quantitatively evaluate the realism and diversity of our synthesized motion compared to ground truth and existing methods. Both quantitative and qualitative results demonstrate the superior performance of our approach, which is capable of generating stable, contact-aware motion sequences from texts. 
Additionally, we show the adaptability of our model for applications in human-object interaction synthesis in static scenes.


\section{Related Work}
\label{sec:related_work}

\subsection{Text-driven human motion generation}
Text-driven motion generation synthesizes 3D human motion from textual descriptions. There are two main categories of methods: latent-space~\cite{humanise, language2pose, ghosh2021synthesis, temos, humanml3d, motionclip, motiongpt, humantomato, ude} and diffusion-based~\cite{motiondiffuse, mdm, mld, mofusion, flame} approaches. 
In the former category, Language2Pose~\cite{language2pose} introduced shared embeddings for language and pose, while \cite{ghosh2021synthesis} extended this concept with a two-manifold space design. TEMOS~\cite{temos} used a transformer structure to encode motion into a single embedding to avoid a tedious auto-regressive process. To address fixed-length prediction issues, \cite{humanml3d} introduced a text2length module for variable-length motion generation.
A recent straightforward text-to-motion framework~\cite{t2mgpt} overcomes these challenges by mapping motion to a discrete latent space and aligning motion and text embeddings with a GPT-like model. Corrupted input training strategies and casual self-attention are adopted to ensure variable motion length.

Regarding diffusion models~\cite{diffusionmodel, stablediffusion, imagen, dalle2}, initially developed for generating complex images, they have demonstrated their effectiveness in various domains, including conditional motion synthesis. MotionDiffuse~\cite{motiondiffuse} and MDM~\cite{mdm} were among the first to employ diffusion models in this context. However, diffusion models are associated with significant computational overhead, intricate sampling procedures, and the need for carefully designed motion noises. An attempt to address motion noise and temporal inconsistency challenges was made by \cite{mld} by projecting raw motion data into a latent space. While effective, the majority of diffusion-based methods exhibited limited improvements in performance and efficiency compared to traditional approaches~\cite{t2mgpt}. 
While prior methods aim to explore the structure design, they neglect the impact of contact when generating motion from text. Our research pioneers the consideration of human-object contact in text-driven motion generation, introducing a novel contact-aware text-to-motion dataset and approach to fill this gap.

\begin{table*}[t]
  \centering
  \caption{Comparison of various text-to-motion datasets. ``Text: Body part'' and ``Text: Object'' refer to the provision of information \textbf{in textual descriptions} about human body parts and the objects involved in contact within each sequence. ``Text: Per-frame'' denotes the precise alignment of text annotations with the motion intervals in the sequence. ``Contact'' indicates the availability of accurate contact labels. ``Real scene'' distinguishes whether the motion was captured in real-world settings or within controlled lab environments. ``GT motion'' means whether the human motion is mocap or pseudo-GT. ``\#Actions'' indicates the number of action categories.}
  \label{tab:datasets}
  \resizebox{\textwidth}{!}{
  \begin{tabular}{l|ccccccc}
    \toprule
    Dataset & Text: Body part & Text: Object & Text: Per-frame & Contact & Real scene & GT motion & \#Actions \\
    \hline
    HumanML3D~\cite{humanml3d}   & \ding{55} & \ding{55} & \ding{55} & \ding{55} & \ding{55} & Mixed & - \\
    KIT-ML~\cite{kitml}          & \ding{55} & \ding{55} & \ding{55} & \ding{55} & \ding{55} & \checkmark & - \\
    BABEL~\cite{babel}           & \ding{55} & \ding{55} & \checkmark & \ding{55} & \ding{55} & \checkmark & 260 \\
    HUMANISE~\cite{humanise}     & \ding{55} & \ding{55} & \ding{55} & \ding{55} & \ding{55} & \checkmark & 4 \\
    HumanAct12~\cite{humanact12} & \ding{55} & \ding{55} & \ding{55} & \ding{55} & \ding{55} & \ding{55} & 12\\
    UESTC~\cite{uestc}           & \ding{55} & \ding{55} & \ding{55} & \ding{55} & \ding{55} & \ding{55} & 40 \\
    NTU-RGB+D~\cite{nturgbd}     & \ding{55} & \ding{55} & \ding{55} & \ding{55} & \ding{55} & \ding{55} & 120 \\
    \hline
    \datasetname                 & \checkmark & \checkmark & \checkmark & \checkmark & \checkmark & \checkmark & 26 \\
    \bottomrule
  \end{tabular}}
\end{table*}

\subsection{Contact-aware motion models}
Human-object interaction, specifically in terms of human-object contact, plays an essential role in understanding and generating natural human actions and behaviors. This aspect has been considered in various domains such as motion reconstruction, motion priors, and motion understanding. In motion reconstruction, many methods~\cite{kevin-physics, contact-dynamics, physcap} address feet-ground contact by training contact detectors, while others leverage physical simulators~\cite{dynamics-regulated, simpoe, diffphy, posetriplet, neuralmocon} and optimization goals inspired by physics principles~\cite{gravicap, shimada-neural}. Motion prior models~\cite{lemo, humor, ma2023grammar} often incorporate human-ground contact beyond the ordinary motion representation to model the transition of human movements. This offers optimization objectives that enhance motion generation and reconstruction toward a physically plausible direction. In motion understanding, some methods~\cite{deco, rich} attempt to infer 3D vertex-level contact on a human body from 2D images, aiming to understand how humans interact with their surroundings. Despite these efforts, the significance of human-object contact has remained relatively unexplored in the context of text-driven human motion generation. To fill this gap, we introduce a novel contact-aware dataset and a simple and effective method to advance research in this domain.

\subsection{Text-to-motion datasets}
Before commencing our work, we require a dataset that encompasses 3D human motions, precise contact labels, and corresponding textual descriptions. Existing text-to-motion datasets have been summarized in Table~\ref{tab:datasets}. Datasets like HumanAct12~\cite{humanact12}, UESTC~\cite{uestc}, and NTU-RGB+D~\cite{nturgbd} cover a wide range of daily actions but offer only basic action labels. HumanML3D~\cite{humanml3d}, KIT-ML~\cite{kitml}, and BABEL~\cite{babel} are well-known text-to-motion datasets, featuring extensive mocap motions. BABEL~\cite{babel}, in particular, includes frame-level textual descriptions. However, these datasets capture data in controlled lab settings and primarily focus on poses and self-contact, often neglecting interactions with other objects. The HUMANISE~\cite{humanise} dataset, while providing 3D scene meshes, is limited to just four indoor actions, and its motion data lacks fine-grained contact information between humans and objects. To address these limitations, we have developed a novel text-to-motion (T2M) dataset based on the RICH~\cite{rich} dataset, including various motion dynamics captured from daily actions. This new dataset contains motion data captured in real-world scenes, diverse indoor and outdoor actions, and accurate vertex- and joint-level contact labels. Furthermore, we have enriched it with frame-level textual annotations that specify interaction between body parts and objects.

\subsection{Human-object interaction generation} 
Human-object interaction (HOI) generation ~\cite{pmp, interdiff, omomo, cghoi, hoidiff} aims to produce a holistic human and object motion sequence conditioned on the object's geometry, sparse object motion evidence, and optionally a textual description. Different from HOI, our task focuses on modeling the alignment between textual descriptions and human motion, 
particularly the varied interactions between human body parts and objects as detailed in the text. Furthermore, our method exhibits versatility in adapting to HOI generation in static scenes. Through the integration of a conditioning module for object geometry, our approach adeptly generates natural and physically plausible motion sequences that interact with the given static objects.

\begin{figure*}
    \centering
    \includegraphics[width=\linewidth]{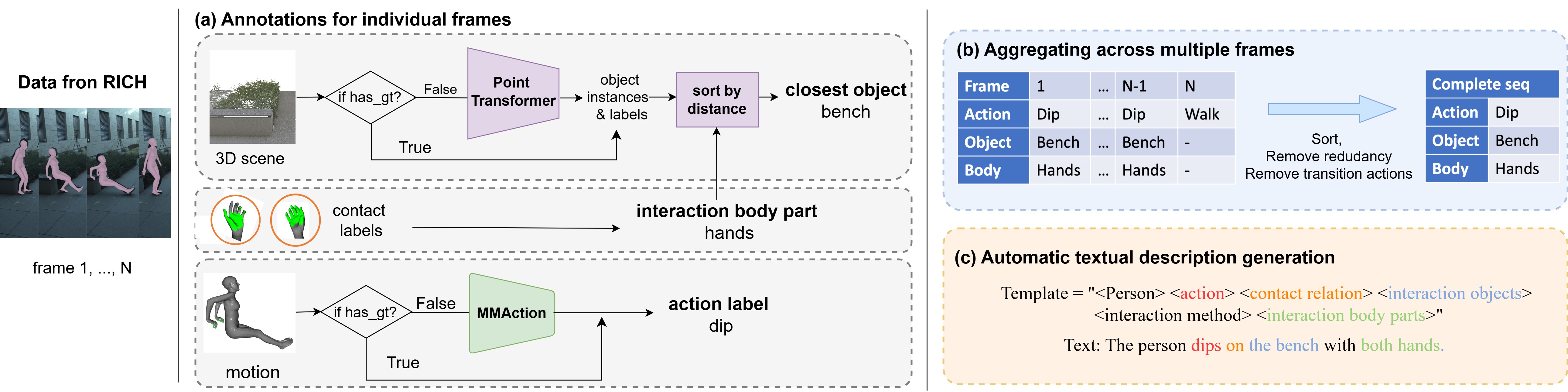}
    \caption{Text annotation pipeline. When presented with a sequence of paired (motion, contact) data and its corresponding 3D scene mesh from RICH~\cite{rich} dataset, our annotation pipeline produces a templated description containing information about actions and interaction details by (a) generating annotations for individual frames, (b) aggregating across mulitple frames, and finally (c) automatically generating textual descriptions.}
    \label{fig:text_anno_pipeline}
\end{figure*}

\section{\datasetname Dataset}
To advance research of text-driven 3D interactive human motion generation, we create a comprehensive dataset \datasetname. This dataset is constructed upon the RICH dataset~\cite{rich}, which provides high-quality motion capture data for various human movements. We automatically generate detailed, interaction-aware textual descriptions with minimal manual effort. The resulting dataset comprises 8,566 motion sequences. Each sequence comes with precise vertex-level contact labels and interaction-aware natural language descriptions that detail the interactions between various body parts and different objects. Direct comparisons to the existing text-to-motion datasets can be found in Table~\ref{tab:datasets}.

As shown in Fig.~\ref{fig:text_anno_pipeline}, our text annotation pipeline (Section~\ref{sec:text_generation}) produces intricate interaction descriptions and eliminates the need for time-consuming manual labeling. We also carefully sample and process the motion and contact data for varying lengths and multi-action scenarios (Section~\ref{sec:data_sampling}). Further details on motion and contact representation can be found in Section~\ref{sec:motion_contact_repr}.

\subsection{Textual Description Generation}\label{sec:text_generation}
In this process, when presented with a sequence of paired (motion, contact) data and its corresponding 3D scene mesh, our annotation pipeline produces a templated description containing information about actions and interaction details with minimal manual effort, as shown in Fig.~\ref{fig:text_anno_pipeline}.

\subsubsection{Annotations for individual frames}
The annotation begins with the 3D segmentation of objects based on motion and contact data, in conjunction with the 3D scene mesh. This segmentation is achieved either through the ground truth annotations from the RICH dataset~\cite{rich} or by employing a pretrained Point Transformer model~\cite{pointtrans}. Subsequently, interaction object labels and corresponding interaction body parts are identified by analyzing the sorted distances between the contacted vertices and each object within the scene. Finally, the action label for each frame is extracted using MMAction2~\cite{mmaction}, unless these labels are already available from the RICH dataset~\cite{rich}.

\subsubsection{Aggregating across multiple frames}
To generate textual descriptions for an entire sequence, we aggregate annotations from multiple individual frames using action ordering and filtering. Initially, all actions are sorted by their time steps, and a unique set of actions for each sequence is created by removing redundant entries. These action types are then categorized into two groups: torso actions and upper-body actions. If a clip contains an excessive number of actions—such as more than two torso actions or more than three upper-body actions—we exclude these clips to maintain a more concise and focused dataset.
Next, labels for actions with a very brief duration (fewer than five frames) are removed, although their motion data is retained in the clips for transitional purposes. Following the filtering process, we order the actions for the entire sequence. In cases where multiple actions occur simultaneously with significant overlap, a random ordering is assigned to these actions.

\subsubsection{Final automatic textual description generation}
For every clip, when provided with a set of labels, we design a template to automatically generate textual descriptions following~\cite{humanise}. The template follows this structure: ``$<$Person$>$ $<$action$>$ $<$contact relation$>$ $<$interaction objects$>$ $<$interaction method$>$ $<$interaction body parts$>$''. The order may vary slightly to align with typical expressions in natural language. In cases of multiple actions, this pattern is repeated and linked together using terms such as ``and''.

\subsection{Data Sampling}\label{sec:data_sampling}
Before the textual annotation process, we must decide what kind of motion data will be included in our \datasetname dataset. To achieve this, we organize all motion sequences based on their action types and scenes. Subsequently, sequences lacking meaningful actions are excluded. The remaining data is then divided into the train, test, and validation sets, approximately at a ratio of 6:1:1. To accommodate variable length conditions in the text-to-motion task, we segment all the data into clips of various lengths, specifically 64, 128, and 192 frames. The quantity of them maintains an approximate ratio of 1:0.8:0.6 in each split. The aforementioned textual annotation process is subsequently applied to each clip to generate the final descriptions. Furthermore, since the quantity of different action types may vary, we adjust the sampling probability and overlap size to ensure a balanced ratio.

\subsection{Motion and Contact Representation}\label{sec:motion_contact_repr}
The original human motion data is represented using body model SMPL-X~\cite{smplx}. This model allows for the calculation of the positions of 10,475 vertices and 54 joints, using parameters for pose ($\theta$), shape ($\beta$), and facial expression ($\phi$). From these, we select the positions of the first 22 joints and compute 263-dimensional motion features following~\cite{humanml3d}.
As for contact labels, binary labels are provided to indicate whether a vertex is in contact or not, for all 10,475 vertices. To maintain consistency with the motion features, we convert them into 22 joint-level contact labels using the regressor provided by the SMPL-X body model~\cite{smplx}.

\begin{figure*}
    \centering
    \includegraphics[width=\linewidth]{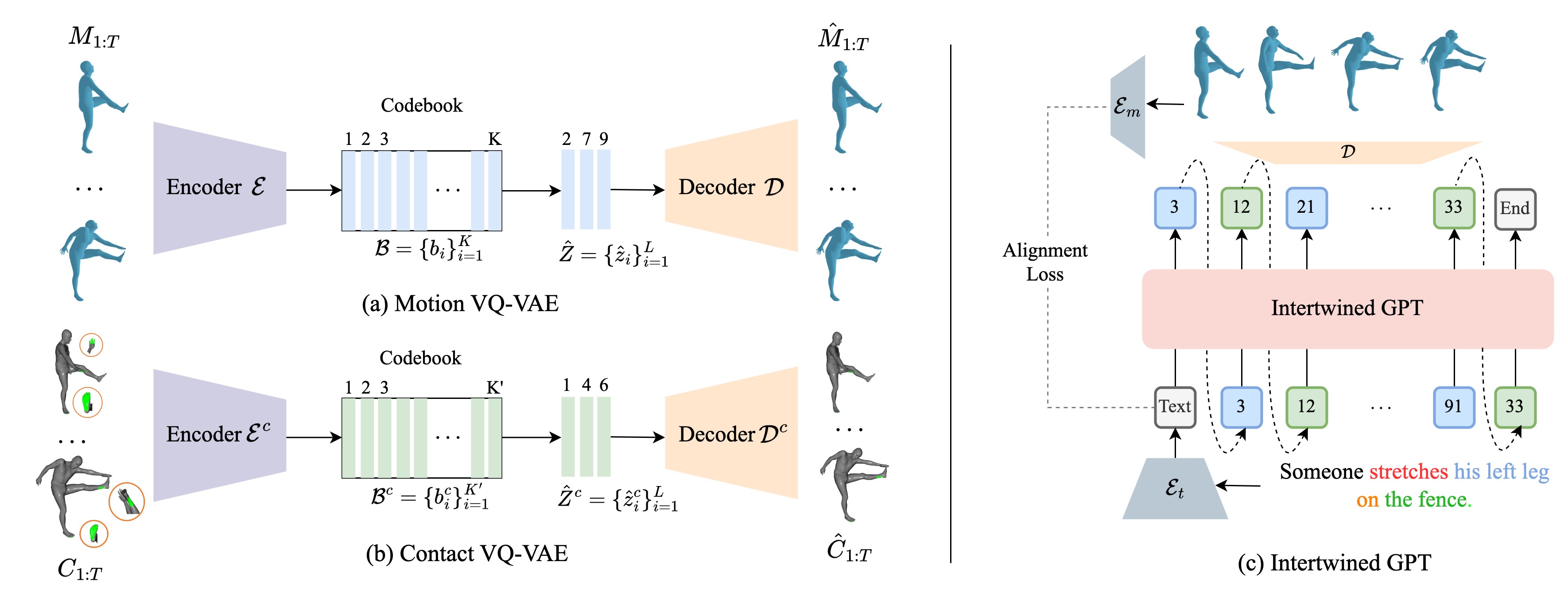}
    \caption{Architecture of our approach \modelname for text-driven 3D interactive human motion synthesis. Our model consists of independent (a) Motion VQ-VAE and (b) Contact VQ-VAE to encode the motion and contact modalities into distinct latent spaces. Subsequently, we autoregressively predict a distribution of motion and contact from the text via (c) the intertwined GPT to explicitly incorporate contact into motion generation. The output from the intertwined GPT is then fed into the learned Motion VQ-VAE decoder $\mathcal{D}$ to yield a sequence of 3D poses with physically plausible interactions. Additionally, the text embedding is extracted from our pretrained text encoder $\mathcal{E}_t$, with an alignment loss ensuring the consistency between interactive text embeddings and the generated poses. $\mathcal{E}_m$ is the movement encoder pretrained with the text encoder $\mathcal{E}_t$ to calculate the alignment loss.}
    \label{fig:architecture}
\end{figure*}

\section{Method}

We aim to generate contact-aware motion sequences conditioned on a textual description that encompasses the interaction details such as interaction body parts and objects. Fig.~\ref{fig:architecture} illustrates the overall method. Concretely, two independent VQ-VAEs are used to encode the motion and contact modalities into distinct yet complementary latent spaces, capturing intricate representations of each modality (Section~\ref{sec:gpt}). Then, we propose an intertwined GPT architecture that generates motion and contact in a twisted auto-regressive way from the text (Section~\ref{sec:gpt}) to explicitly incorporate contact into motion generation. Subsequently, the output from intertwined GPT is fed into the learned Motion VQ-VAE decoder to yield a sequence of 3D poses with physically plausible interactions. Lastly, a pre-trained text-motion model is introduced to effectively decompose interactive textual descriptions and align them with the generated motion sequence by alignment loss (Section~\ref{sec:text_encoder}).

\subsection{Motion and Contact VQ-VAEs}\label{sec:vqvae}
Considering continuous motion features and binary contact labels as distinct modalities, we adopt two separate VQ-VAE models~\cite{vqvae} to encode motion and contact modalities into distinct yet complementary latent spaces. The key insight for the separated modeling is to capture the inherently diverse and sophisticated patterns of each modality and give each aspect full expressiveness to facilitate the subsequent generative process.

Specifically, each VQ-VAE model consists of an encoder $\mathcal{E}$, a decoder $\mathcal{D}$, and a codebook $\mathcal{B}$. The encoder $\mathcal{E}$ first applies a 1D convolution (Conv1D) layer to the motion sequence $M = [m_1, m_2, \cdots, m_T]$ or contact sequence $C = [c_1, c_2, \cdots, c_T]$ along the temporal dimension to compress by a ratio of $l$. A series of Conv1D layers and residual blocks follow to convert the sequence of data into latent feature $Z = [z_1, z_2, \cdots, z_{L}]$ of length $L=T//l$ and dimension $D$. Then the quantitized latent feature $\hat{Z} = [\hat{z}_1, \hat{z}_2, \cdots, \hat{z}_L]$ is collected by fetching the most similar discrete elements in a learned codebook $\mathcal{B} = \{b_k\}_{k=1}^K$ by $\hat{z}_i = \arg\min_k ||z_i - b_k||^2$. Finally, the reconstructed motion $\hat{M} = [\hat{m}_1, \hat{m}_2, \cdots, \hat{m}_T]$ or contact $\hat{C} = [\hat{c}_1, \hat{c}_2, \cdots, \hat{c}_T]$ sequence is recovered by decoder $\mathcal{D}$. 

For each VQ-VAE, we train the encoder, decoder, and codebook simultaneously with the following three components: reconstruction loss $\mathcal{L}_{re}$, embedding loss $\mathcal{L}_e$, and commitment loss $\mathcal{L}_c$. Both Motion VQ-VAE and Contact VQ-VAE are trained independently and share the same loss functions for embedding loss and commitment loss, as expressed in the following:
\begin{equation}
    \mathcal{L}_{e} = || SG(Z) - \hat{Z}||_2, \ \ \mathcal{L}_c = || Z - SG(\hat{Z})||_2,
\end{equation}
where $SG$ is stop-gradient. $Z$ and $\hat{Z}$ are the ground truth latent feature and the predicted one, respectively. 

For the reconstruction loss $\mathcal{L}_{re}$, we employ Binary Cross-Entropy (BCE) loss to the output logits $\hat{C}$ after softmax operation:
\begin{equation}
    \mathcal{L}_{re} = - \frac{1}{T*J} \sum_{t,j} ( (1-C_{t,j}) \log(1 - \hat{C}_{t,j}) + C_{t,j}\log(\hat{C}_{t,j})),
\end{equation}
where $T$ is the total length of the sequence, and $J$ is the number of joints. For Motion VQ-VAE, we apply $l1$-smooth loss to both motion feature and motion velocity following \cite{t2mgpt}, which is omitted here.

The loss for each VQ-VAE is summarized as follows. Note that $\alpha$ and $\beta$ are the hyper-parameters for balancing loss items.
\begin{equation}
    \mathcal{L}_{vq} = \mathcal{L}_{re} + \alpha\mathcal{L}_{e} + \beta\mathcal{L}_{c}.
\end{equation}

\subsubsection{Implementation details}
Motion VQVAE and Contact VQVAE both adopt a series of Conv1D and residual blocks with a downsampling rate set to $l=4$. During training, we employ Exponential Moving Average (EMA), codebook reset~\cite{vqvae, t2mgpt} training strategies to prevent codebook collapse. 
For VQ-VAEs, we adopt a codebooks zie of 1024 for Motion VQ-VAE and 512 for Contact VQ-VAE. 
Moreover, to explore their impact, we conducted an ablation study on the codebook sizes of both VQ-VAEs. The details are presented in Section~\ref{sec:ablation}. 

\subsection{Intertwined GPT}\label{sec:gpt}
\subsubsection{Intertwined motion and contact GPT}
After we learn the separate quantized codebooks, any motion or contact sequence ($C$ or $M$ respectively) of length $T$ can be represented as a quantitized latent feature $\hat{Z}$ of $L=T//l$ units and yields the index sequence $S$ of the same length through quantization. 
With the quantized motion and contact representation, we now consider generating interactive motion from text. Typically, it is formulated as a next-index prediction problem~\cite{t2mgpt}. Given indices for $i-1$ units $S = [s_1, s_2, \cdots, s_{i-1}]$ and the input text embedding (Section~\ref{sec:text_encoder}), the goal is to predict the next index $s_i$, and the learned Motion VQ-VAE decoder $\mathcal{D}$ is then used to reconstruct the motion sequence $\hat{M}$ from the generated indices. 

\begin{figure}
    \centering
    \includegraphics[width=\linewidth]{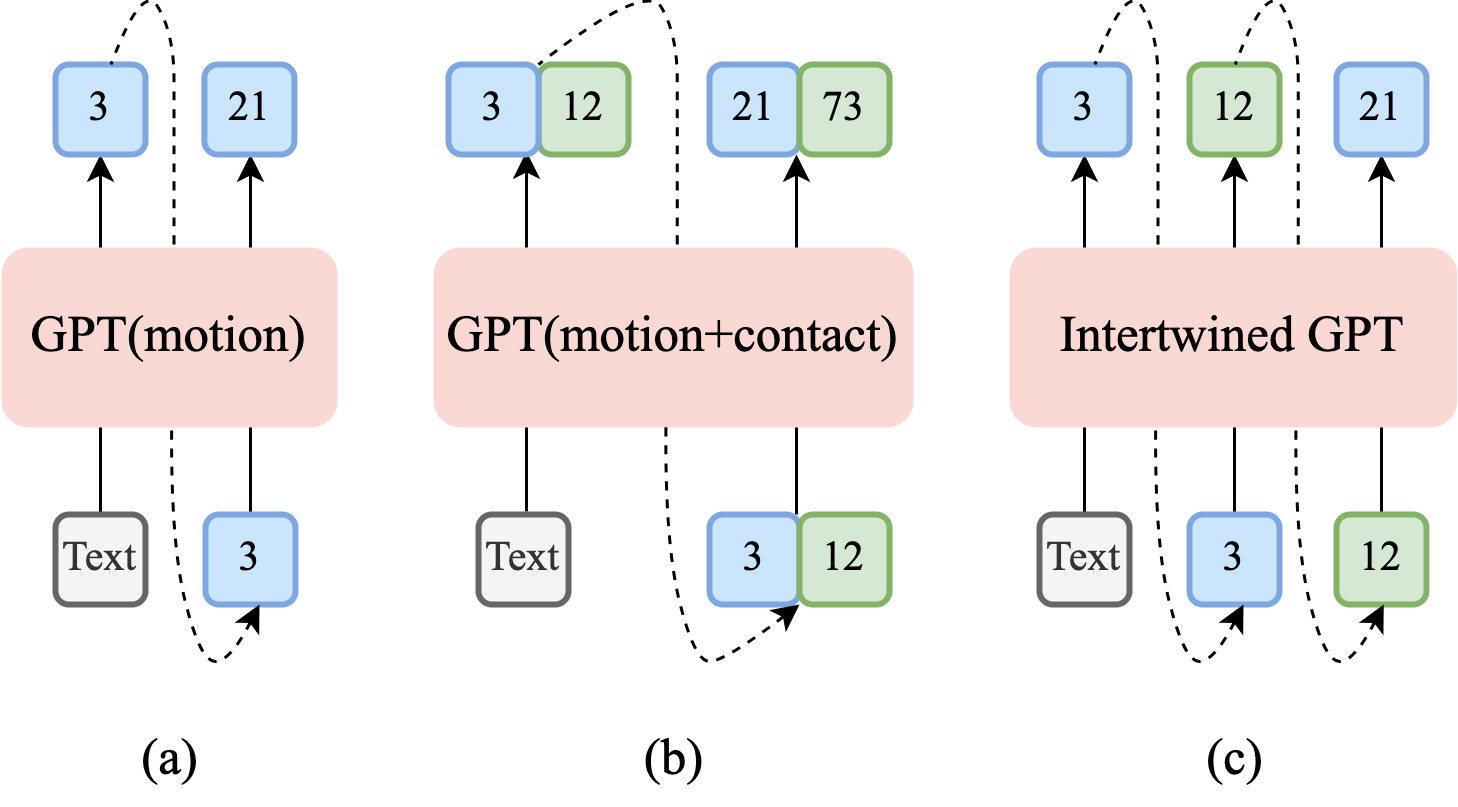}
    \caption{Architecture comparison between Intertwined GPT and ordinary GPTs. Standard GPT either predicts only motion (a), or parallel motion and contact tokens sequentially (b). Instead, Intertwined GPT (c) predicts them in a cross-conditioned manner, allowing contact-aware motion generation}
    \label{fig:gpt_comparison}
\end{figure}

To generate interaction-aware motion sequences, we introduce an interleaved approach that seamlessly integrates contact token indices $S^c = [s^c_1, s^c_2, \cdots, S^c_{L}]$ derived from the same video clip into the generation process. 
Instead of treating contact tokens as additional input features, we capitalize on the temporal correlations between motion and contact sequences by inputting intertwined motion and contact sequences $X = [s_1, s^c_1, s_2, s^c_2, \cdots, s_{i-1}, s^c_{i-1}]$, where the motion and contact indices for each frame are positioned side by side along the temporal dimension. This allows the model to learn to separate the two types of token indices and predict the next motion index $s_i$ or contact index $s_i^c$ conditioned on both past motion and contact indices. The joint probability of the whole sequence is modeled as follows:  
\begin{equation}
\begin{split}
    p(S_{1:L}, S^c_{1:L} | Z_t) = \prod_{i=1}^{L} & p(s_{i} | s_{<i}, s^c_{<i}, Z_t) p(s^c_{i} | s_{\leq i}, s^c_{<i}, Z_t),  \\
\end{split}
\end{equation}
where $Z_t$ is the text embedding extracted by our interaction-aware text encoder introduced in Section~\ref{sec:text_encoder}. The architecture comparison between our Intertwined GPT and ordinary GPTs are presented in Fig.~\ref{fig:gpt_comparison}.

During inference, the motion and contact sequences will be generated in an autoregressive manner conditioned on text embedding $Z_t$. There will be two cases to consider: 1) predicting the next-unit motion index $s_{i}$ based on motion and contact indices of past units $s_{<i}, s^c_{<i}$; 2) inferring the current-unit contact index $s^c_{i}$ based on the current-unit motion index $s_i$ and all previous indices $s_{<i}, s^c_{<i}$. For each step, there are no specific mode flags to indicate the case. This design, combined with the multi-modal corruption training strategy introduced later, effectively leverages the complementary nature of motion and contact within the same frame and the continuity across different frames, leading to more stable and temporally consistent outcomes. 

\subsubsection{Multi-modal corruption training strategy}
Generally, to reduce the discrepancy between training and testing, a common practice is to randomly corrupt a portion of the input sequence during training to enhance model robustness. In the case of simultaneously inputting motion and contact indices, a reasonable approach would be to randomly replace motion and contact indices $[s_i, s^c_i]$ for certain frames with random values. However, this practice might cause the model to neglect the scenario where the current-unit contact index $s_i^c$ is generated based on the history of indices and the current-unit motion index $S = [s_1, s^c_1, \cdots, s_i]$. Consequently, the connection and interaction between motion and contact may diminish. Therefore, we adopt a mixed-random strategy, treating each token index equally and randomly corrupting a portion during training.

\subsubsection{Objective functions}
Our objectives encompass two aspects: the reconstruction capability for corrupted indices and the next motion/contact index, as well as the alignment between text and motion. For the former, we employ Cross-Entropy (CE) loss. Regarding the latter, we calculated the contrastive loss between the embeddings of text and the recovered motion sequence. First, the motion sequence is recovered from the predicted indices via the Motion VQ-VAE decoder. Then the motion embedding is extracted using the pre-trained motion model mentioned in Section~\ref{sec:text_encoder}. Finally, we calculate contrastive loss with the input text embedding.

\subsection{Interaction-aware Text Encoder}\label{sec:text_encoder}
For text embedding, recent methods~\cite{motionclip,motiongpt,t2mgpt} have utilized language model text encoders like CLIP~\cite{clip} and sBERT~\cite{sbert}. While these models exhibit good generalization, they are not specifically finetuned on datasets containing descriptions of interactive motions. We observed that CLIP struggles to fully comprehend textual information with interactions, assigning high similarity to descriptions of different interactions but similar actions. For instance, descriptions like ``\textit{push up on the ground}’’ and ``\textit{push up on the fence}’’ might differ by only one word, leading to minimal distinction in text embeddings for CLIP. However, in motion, these actions exhibit significant differences in global orientation and position. This discrepancy significantly affects retrieval results. 

Inspired by \cite{tmr}, we employ a straightforward pretrained text-motion matching model designed to align interaction components between text and motion. Our text-motion matching model architecture is straightforward, comprising a text encoder to extract textual embeddings, and a movement encoder to subsequently extract interactive motion embeddings. The embeddings from both modalities are aligned through contrastive loss. After pretraining, the frozen text encoder is used to extract text embeddings as inputs for the generative model. 

\subsubsection{Architecture and Pretraining}
The interaction-aware text encoder is trained alongside a movement encoder, aiming at generating text and motion embeddings that seamlessly align. Each encoder comprises multiple convolutional and MLP layers. Specifically, the interaction-aware text encoder is designed to encode detailed textual descriptions, incorporating labels for interacting objects and body parts. On the other hand, the movement encoder specializes in encoding motion sequences into distinct latent spaces. Movement encoder consists of two parts: the first part is trained with reconstruction loss to extract motion embeddings, while the second part, comprising additional MLP layers, is trained with contrastive loss to facilitate motion-text alignment.

The model undergoes pretraining using the training set of our \datasetname dataset. The initial training phase involves independently training the first part of the movement encoder with the reconstruction loss. Subsequently, during joint training of the text encoder and movement encoder with contrastive loss for alignment, the first part of the movement encoder remains frozen.

\begin{table*}[t]
  \centering
  \caption{Comparison with the state-of-the-art methods on our \datasetname test set. * denotes that we replace the motion representation in the method with a concatenated representation of motion and contact states. \textcolor{red}{Red} and \textcolor{blue}{blue} highlight the best and the second best results. Following~\cite{humanml3d}, we conduct the evaluation process 20 times and report the average results with a 95\% confidence level.}
  \label{tab:main_results}
  
  \scalebox{1}{
  \begin{tabular}{l c c c c c c c}
    \toprule
    \multirow{2}{*}{Methods}  & \multicolumn{3}{c}{R-Precision $\uparrow$} & \multirow{2}{*}{FID $\downarrow$} & \multirow{2}{*}{MM-Dist $\downarrow$} & \multirow{2}{*}{Diversity $\uparrow$} & \multirow{2}{*}{MModality $\uparrow$}\\
    \cline{2-4}
    ~ & Top-1 & Top-2 & Top-3 \\
    \midrule
    Real motion & $0.492^{\pm 0.003}$ & $0.746^{\pm 0.002}$ & $0.870^{\pm 0.002}$ & $0.000^{\pm 0.000}$ & $3.020^{\pm 0.003}$ & $12.868^{\pm 0.111}$ & - \\
    Motion VQVAE & $0.471^{\pm 0.003}$ & $0.722^{\pm 0.003}$ & $0.847^{\pm 0.003}$ & $0.443^{0.004}$ & $3.382^{\pm 0.004}$ & $12.933^{\pm 0.097}$ & -\\
    \midrule
    MDM~\cite{mdm} & $0.360^{\pm 0.004}$ & $0.561^{\pm 0.001}$ & $0.673^{\pm 0.005}$ & $0.949^{\pm 0.037}$ & $4.873^{\pm 0.032}$ & $13.077^{\pm 0.085}$ & ${2.358}^{\pm 0.139}$ \\
    T2M-GPT~\cite{t2mgpt} & $0.462^{\pm 0.004}$ & $0.697^{\pm 0.005}$ & $0.816^{\pm 0.003}$ & $1.195^{\pm 0.034}$ & $3.385^{\pm 0.018}$ & $13.089^{\pm 0.119}$ & ${2.396}^{\pm 0.043}$ \\
    MLD~\cite{mld} & $\textcolor{blue}{0.489}^{\pm 0.003}$ & $\textcolor{blue}{0.741}^{\pm 0.002}$ & $\textcolor{blue}{0.866}^{\pm 0.002}$ & $\textcolor{blue}{0.771}^{\pm 0.012}$ & $\textcolor{blue}{2.904}^{\pm 0.014}$ & $\textcolor{red}{13.396}^{\pm 0.101}$ & $2.095^{\pm 0.082}$ \\
    \midrule
    MDM* \cite{mdm} & \et{0.102}{0.003} & \et{0.179}{0.004} & \et{0.247}{0.006} & \et{20.620}{0.355} & \et{10.144}{0.047} & \et{11.631}{0.182} & \etr{6.589}{0.001} \\
    T2M-GPT*\cite{t2mgpt} & \et{0.466}{0.003} & \et{0.696}{0.004} & \et{0.808}{0.004} & \et{0.895}{0.025} & \et{3.512}{0.019} & \et{12.784}{0.110} & \et{2.607}{0.095} \\
    MLD*\cite{mld} & \et{0.482}{0.003} & \et{0.718}{0.003} & \et{0.832}{0.002} & \et{0.868}{0.015} & \et{3.225}{0.016} & \et{13.068}{0.098} & \etbb{2.622}{0.074}\\
    \midrule
    Ours & $\textcolor{red}{0.529}^{\pm 0.003}$ & $\textcolor{red}{0.766}^{\pm 0.002}$ & $\textcolor{red}{0.875}^{\pm 0.002}$ & $\textcolor{red}{0.622}^{\pm 0.012}$ & $\textcolor{red}{2.764}^{\pm 0.016}$ & $\textcolor{blue}{13.188}^{\pm 0.139}$ & $2.162^{\pm 0.060}$ \\
    \bottomrule
  \end{tabular}
  }
\end{table*}

\subsubsection{Text encoder in the Intertwined GPT}
While training the intertwined GPT, the text encoder remains frozen and produce text embeddings as input for the intertwined GPT. The alignment loss, using the same contrastive loss~\cite{ctloss}, is employed between the text embedding and the motion embedding to finetune the intertwined GPT. This motion embedding is generated by the frozen movement encoder based on the prediction from the intertwined GPT.

To calculate contrastive loss $L$, given the motion embedding $Z$ and a text embedding $Z_t$, we can calculate the pairwise euclidean distance between $Z$ and $Z_t$ as follows.
\begin{equation}
 D(Z_t, Z) = || Z_t - Z ||_2.
\end{equation}
Let $Y$ be a binary label to indicate whether the text and motion embeddings belong to the same sequence. if text and motion embeddings belong to the same distance, $Y=0$, otherwise, $Y=1$. Then the loss function can be formulated as
\begin{equation}
\begin{split}
 & \mathcal{L} = (1-Y) \mathcal{L}_s (Z_t, Z) + Y \mathcal{L}_d (Z_t, Z), \\
 & \mathcal{L}_s = D(Z_t, Z), \mathcal{L}_d = D(Z_t, Z).
\end{split}
\end{equation}
Where $\mathcal{L}_s$ is a positive loss function for a pair of text and motion embeddings that belong to the same sequence, and $\mathcal{L}_d$ is the negative loss function for the data that belongs to different sequences.
During implementation, we randomly shifted the data in a batch by an offset, and calculate the negative loss between the text embedding $Z_t$ and the randomly shifted motion embedding $Z'$. A margin $m = 10$ is adopted to control the magnitude of negative loss as follows,
\begin{equation}
\mathcal{L}_d = || max(0, m - D(Z_t, Z')) ||_2.
\end{equation}

\section{Experiment}

\subsection{Experimental Settings}

\subsubsection{Comparison methods}
We evaluate our method \modelname and three state-of-the-art text-to-motion methods on our \datasetname dataset, including one latent-space method T2M-GPT~\cite{t2mgpt} and two diffusion-based methods MDM~\cite{mdm}, MLD~\cite{mld}. 
For a fair comparison, we modified the above methods by directly concatenating the contact features with the motion inputs, denoting MDM*, T2M-GPT*, MLD*.
All the methods are trained on our \datasetname training set and evaluated on the test set. 

\subsubsection{Evaluation metrics}
Following \cite{humanml3d}, we pretrained an evaluation model on the \datasetname training set. Motion and text embeddings are derived from it and assessed using five commonly used metrics. \textit{Frechet Inception Distance (FID)} measures the distance between the generated motion and the ground truth, reflecting the quality of the generated motion sequences. \textit{R-Precision} assesses the alignment quality between text and motion embeddings, reporting Top-1, Top-2, and Top-3 accuracy, with a batch size of 32 for calculation. \textit{Multimodal Distance (MM-Dist)} computes the average distance between each text and motion embedding pair, while \textit{Multimodality (MModality)} calculates the pairwise distance between 10 pairs of motion sequences generated from the same text description. Lastly, for \textit{Diversity}, we randomly sample 300 pairs of motion sequences from the entire set and compute the average distance.

\subsubsection{Implementation details}
We implement comparison methods on our \datasetname dataset using their official repositories. For the VQ-VAE models employed in both T2M-GPT~\cite{t2mgpt} and our approach, we set the downsampling rate $l$ to $4$. We train 300,000 iterations for all VQ-VAEs and 100,000 iterations for GPT, using a learning rate of $2e^{-4}$ and a batch size of $16$. For the diffusion models, MDM~\cite{mdm} and MLD~\cite{mld}, default settings are adopted. For MLD~\cite{mld}, we conduct a training of 1500 epochs for the VAE and an additional 1500 epochs for the diffusion model. For their variants, we replace the motion input with the concatenation of both motion and contact features. 

\begin{figure*}[t]
    \centering
    \includegraphics[width=\linewidth]{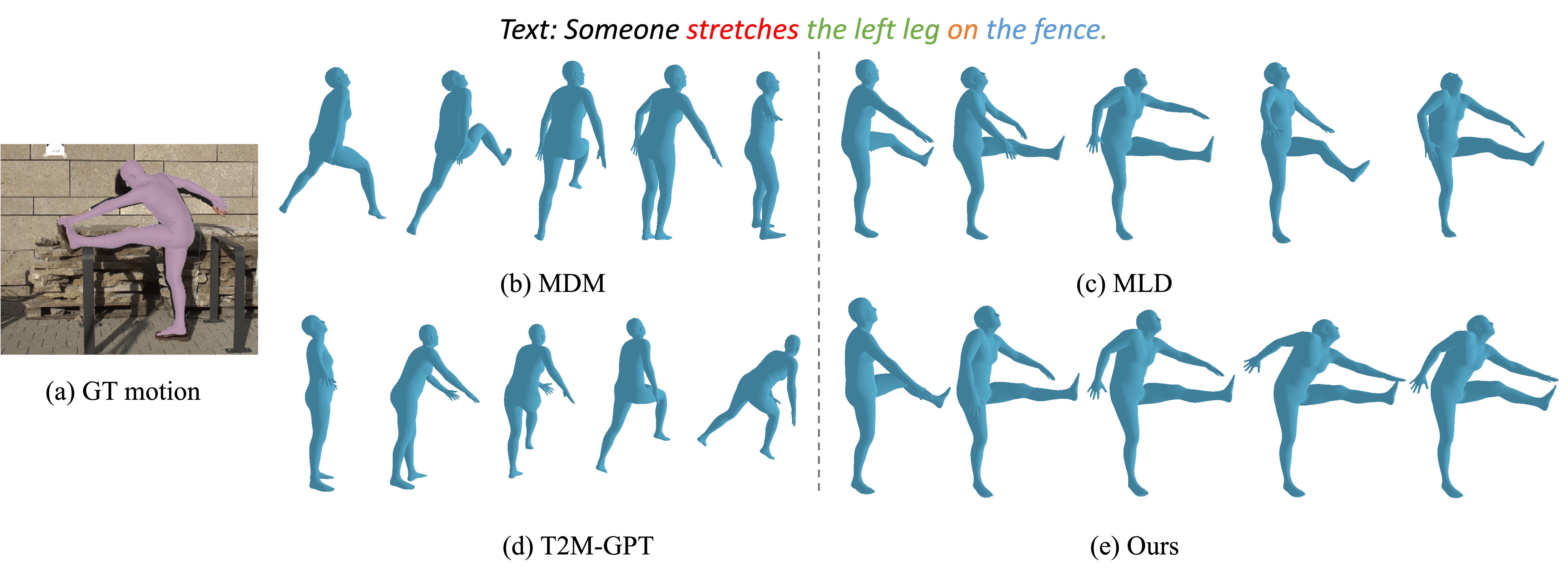}
    \includegraphics[width=\linewidth]{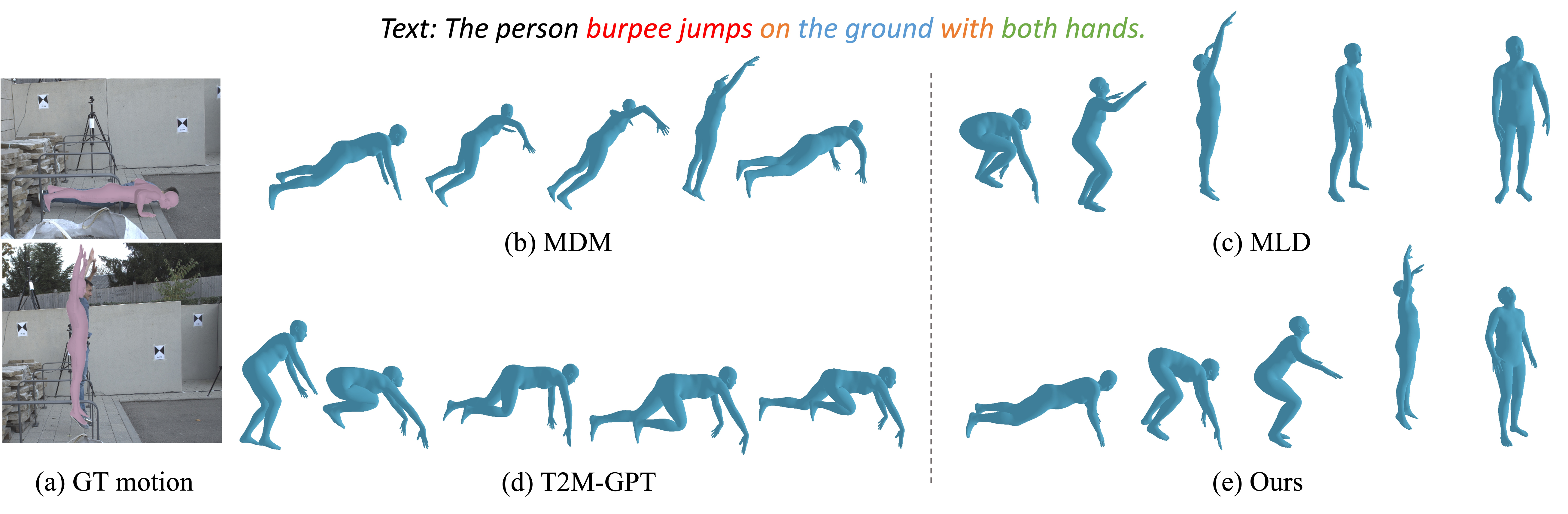}
    \caption{Qualitative comparison with the state-of-the-art methods on \datasetname test set. We compare (e) our method with (b) MDM~\cite{mdm}, (c) MLD~\cite{mld}, (d) T2M-GPT~\cite{t2mgpt}. Part of (a) ground-truth motion is provided for reference.}
    \label{fig:visual_res}
\end{figure*}

\subsection{Main Results}
\subsubsection{Quantitative results}
Table~\ref{tab:main_results} displays the numerical results on our \datasetname test set. We compare our model with three state-of-the-art methods and their modified versions: T2M-GPT~\cite{t2mgpt}, MDM~\cite{mdm} and MLD~\cite{mld}. The results demonstrate that our method outperforms the baselines and their variants in both motion generation and text-motion consistency. 
Particularly, our method outperforms others significantly in the FID metric, indicating high-quality interactive motion generation. In terms of text-motion consistency, we outperform other methods significantly in R precision metrics and multimodal distance, and achieve better outcomes in Multimodality compared to MLD~\cite{mld}.
Additionally, compared with the original versions, T2M-GPT* performs better with integrated contact maps, while MLD* experiences a slight degradation. Notably, MDM* collapses, due to the diffusion model’s limited suitability for handling motion and interaction data with diverse distributions and data formats. MLD* avoids this issue by leveraging VAE to extract powerful latent before employing diffusion model.

\subsubsection{Qualitative results}
Fig.~\ref{fig:visual_res} shows the visual results on our \datasetname test set. We compare our results with T2M-GPT~\cite{t2mgpt}, MDM~\cite{mdm}, and MLD~\cite{mld}. The generation results of first case show that our method can accurately understand all the factors in the text, including action, interaction object (fence), and interaction body part (left leg). The other methods exhibit mismatches between the action \textit{“climb over the fence”} and \textit{“stretch on the fence”}, resulting in a wrong action type. In contrast, our method can distinguish the two similar poses by the learned contact data, and respond to the object specified in the textual input. As depicted in Fig.~\ref{fig:visual_res}(c), MLD accurately identifies the action “\textit{stretches the left leg}”, but it lacks effective interaction with the object “\textit{fence}”. The left leg should remain stationary, yet MLD's result displays noticeable jittering in both upward and downward directions, suggesting an inaccurate perception of the object's position. In contrast, our approach maintains the left leg in stable and realistic positions.

In the second case, our method accurately executes the burpee jump action, maintaining coherent ground contacts. T2M-GPT~\cite{t2mgpt} and MDM~\cite{mdm} fail to complete the action, with MDM~\cite{mdm} exhibiting significant jittering and implausible motion. While MLD~\cite{mld} performs the action correctly, it shows skating and jittering on the feet. In contrast, our method completes the entire burpee jump action on the ground with accurate body-ground interactions.

\begin{table*}[t]
    \centering
    \caption{Ablation study. ``M'': Motion VQ-VAE. ``C'': Contact VQ-VAE. ``iGPT'': intertwined GPT. ``Cr'': multi-modal corruption training strategy. ``$\mathcal{E}_t$'': interaction-aware text encoder pretrained on our \datasetname training set. ``Align'': text-motion alignment loss.}
    \label{tab:ablation_full}
    \scalebox{0.93}{
    \begin{tabular}{c c c c c c c|c c c c c c c}
    \toprule
    \multirow{2}{*}{M} & \multirow{2}{*}{C} & \multirow{2}{*}{GPT} & \multirow{2}{*}{iGPT} & \multirow{2}{*}{Cr} & \multirow{2}{*}{$\mathcal{E}_t$} & \multirow{2}{*}{align} & \multicolumn{3}{c}{R-Precision $\uparrow$} & \multirow{2}{*}{FID $\downarrow$} & \multirow{2}{*}{MM-Dist $\downarrow$} & \multirow{2}{*}{Diversity $\uparrow$} & \multirow{2}{*}{MModality $\uparrow$}\\
    \cline{8-10}
    ~ & ~ & ~ & ~ & ~ & ~ & ~ & Top-1 & Top-2 & Top-3 \\
    \hline
    \checkmark & ~ & \checkmark & ~ & ~ & ~ & ~ & \et{0.462}{0.004} & \et{0.697}{0.005} & \et{0.816}{0.003} & \et{1.195}{0.034} & \et{3.385}{0.018} & \et{13.089}{0.119} & \et{2.396}{0.043} \\
    \checkmark & \checkmark & \checkmark & ~ & ~ &  ~ & ~ & \et{0.455}{0.003} & \et{0.690}{0.003} & \et{0.814}{0.002} & \et{0.833}{0.028} & \et{3.497}{0.014} & \et{13.039}{0.098} & \et{2.708}{0.111} \\
    \checkmark & \checkmark & ~ &\checkmark & ~ & ~ & ~ & \et{0.488}{0.004} & \et{0.726}{0.004} & \et{0.842}{0.003} & \et{0.817}{0.030} & \et{3.202}{0.019} & \et{12.967}{0.139} & \et{2.430}{0.107}\\
    \checkmark & \checkmark & ~ & \checkmark & \checkmark & ~ & ~ & \et{0.484}{0.003} & \et{0.720}{0.004} & \et{0.840}{0.003} & \et{0.785}{0.013} & \et{3.203}{0.021} & \et{13.032}{0.132} & \et{2.359}{0.047} \\
    \checkmark & \checkmark & ~ & \checkmark & \checkmark & \checkmark & ~ & \et{0.514}{0.004} & \et{0.747}{0.003} & \et{0.856}{0.002} & \et{0.632}{0.025} & \et{3.030}{0.022} & \et{12.873}{0.119} & \et{2.838}{0.057} \\
    \checkmark & \checkmark & ~ & \checkmark & \checkmark & \checkmark & \checkmark & \et{0.529}{0.003} & \et{0.766}{0.002} & \et{0.875}{0.002} & \et{0.622}{0.012} & \et{2.764}{0.016} & \et{13.188}{0.139} & \et{2.162}{0.060}\\
    \bottomrule
    \end{tabular}
    }
\end{table*}

\begin{table*}
    \centering
    \caption{Ablation study for motion and contact codebook sizes. ``M'': the codebook size of Motion VQ-VAE. ``C'': the codebook size of Contact VQ-VAE. All the models are trained without alignment loss.}
    \label{tab:ablation_codebook_full}
    \scalebox{1}{
    \begin{tabular}{cc|ccccccc}
    \toprule
    \multirow{2}{*}{M} & \multirow{2}{*}{C} & \multicolumn{3}{c}{R-Precision $\uparrow$} & \multirow{2}{*}{FID $\downarrow$} & \multirow{2}{*}{MM-Dist $\downarrow$} & \multirow{2}{*}{Diversity $\uparrow$} & \multirow{2}{*}{MModality $\uparrow$}\\
    \cline{3-5}
    ~ & ~ & Top-1 & Top-2 & Top-3 \\
    \hline
    $256$ & $512$ & \et{0.479}{0.003} & \et{0.700}{0.003} & \et{0.804}{0.003} & \et{2.521}{0.058} & \et{3.903}{0.027} & \et{13.171}{0.079} & \et{3.736}{0.125}\\
    $512$ & $512$ & \et{0.515}{0.003} & \et{0.761}{0.003} & \et{0.876}{0.002} & \et{0.772}{0.023} & \et{2.831}{0.016} & \et{13.184}{0.108} & \et{2.225}{0.038} \\
    $1024$ & $512$ & \et{0.514}{0.004} & \et{0.747}{0.003} & \et{0.856}{0.002} & \et{0.632}{0.025} & \et{3.030}{0.022} & \et{12.873}{0.119} & \et{2.838}{0.057} \\
    \hline
    $512$ & $256$ & \et{0.517}{0.003} & \et{0.762}{0.003} & \et{0.875}{0.003} & \et{0.727}{0.018} & \et{2.856}{0.019} & \et{13.268}{0.104} & \et{2.122}{0.100}\\
    $512$ & $512$ & \et{0.515}{0.003} & \et{0.761}{0.003} & \et{0.876}{0.002} & \et{0.772}{0.023} & \et{2.831}{0.016} & \et{13.184}{0.108} & \et{2.225}{0.038} \\
    $512$ & $1024$ & \et{0.437}{0.003} & \et{0.643}{0.005} & \et{0.748}{0.004} & \et{4.365}{0.120} & \et{4.513}{0.034} & \et{12.613}{0.135} & \et{4.171}{0.143}\\
    \bottomrule
    \end{tabular}
    }
\end{table*}

\subsection{Ablation Study}\label{sec:ablation}

\subsubsection{Motion and contact modeling}
Table~\ref{tab:ablation_full} presents the performance analysis of various components within our method on the \datasetname dataset. The combination of our Motion VQ-VAE and Contact VQ-VAE, even without the use of the intertwined GPT, demonstrates notable effectiveness compared to a singular Motion VQ-VAE, as evidenced by a significant decrease in FID, approximately 0.36. The multimodality metric also sees an improvement of about 0.31, indicating that integrating contact information can significantly enhance motion quality.

The introduction of the intertwined GPT significantly enhances R-precision, leading to increases of approximately 3.3\%, 3.6\%, and 2.8\% in Top-1, Top-2, and Top-3 accuracy, respectively, thereby demonstrating an improved model capability in text-to-motion retrieval. Additionally, the multimodal distance metric improves by 0.278, reflecting a more realistic motion prediction. This improvement is largely due to the mutually conditioned prediction of motion and contact data within the intertwined GPT module. Furthermore, the implementation of a multimodal corruption strategy results in a substantial decrease in the FID metric, from 0.817 to 0.785. It is important to note that models not utilizing the multimodal corruption strategy employ a standard corruption strategy during training.

Upon further integration of the interaction-aware text encoder, there is a remarkable enhancement in text-motion alignment, with an increase of approximately 3\%, 2.7\%, and 1.6\% in Top-1, Top-2, and Top-3 accuracy, respectively, and a decrease of about 0.153 in FID. This improvement is attributed to our text encoder's enhanced understanding of interactions in textual descriptions.

The incorporation of the alignment loss further enhances the accuracy of motion generation, leading to a notable improvement in both text-motion alignment and motion generation, with particularly significant gains observed in Multimodality and R-precision.

\subsubsection{Impact of codebook sizes}
To explore the impact of the codebook sizes of both Motion and Contact VQ-VAEs, We set one codebook size to a fixed value of $512$ and experiment with the other, varying from $[256, 512, 1024]$. We then compare the results of VQ-VAEs and intertwined GPT trained with these parameters on \datasetname test set. 

Analyzing the outcomes in Table~\ref{tab:ablation_codebook_full}, it becomes evident that employing a larger codebook size for Motion VQ-VAE and a smaller codebook size for Contact VQ-VAE yields superior results. When the contact codebook is fixed at 512, a small motion codebook size (256) results in significantly poorer outcomes, including a notable increase in FID by more than 1.7, and a distinct decrease in R precision, surpassing 5\% in Top-3 accuracy. Similarly, setting the motion codebook size to 512, in cases where the contact codebook size is large (1024), leads to comparable negative effects. Intuitively, motion features, with a higher dimensionality (263 dimensions), are more intricate compared to contact features (22 dimensions), necessitating a larger codebook. 

\subsubsection{Impact of prediction order of intertwined GPT}
The intertwined GPT is engineered to predict both the motion token and the contact token in a mutually conditioned fashion. Initially, the model predicts the motion token, followed by the contact token. To investigate the influence of this prediction order, we have experimented with reversing it. The results are summarized in Table~\ref{tab:ablation_prediction_order}, where it can be seen that even with the reversed order, the intertwined GPT continues to achieve good results on our \datasetname test set.

\begin{table*}[t]
    \centering
    \caption{Ablation study for prediction order of intertwined GPT. This study is conducted on motion and contact encoders with a uniform codebook size of 512.}
    \label{tab:ablation_prediction_order}
    \scalebox{1}{
    \begin{tabular}{c|ccccccc}
    \toprule
    \multirow{2}{*}{Order} & \multicolumn{3}{c}{R-Precision $\uparrow$} & \multirow{2}{*}{FID $\downarrow$} & \multirow{2}{*}{MM-Dist $\downarrow$} & \multirow{2}{*}{Diversity $\uparrow$} & \multirow{2}{*}{MModality $\uparrow$}\\
    \cline{2-4}
    ~ & Top-1 & Top-2 & Top-3 \\
    \hline
    motion $\rightarrow$ contact &  ${0.512}^{\pm 0.003}$ & ${0.755}^{\pm 0.004}$ & ${0.870}^{\pm 0.003}$ & ${0.743}^{\pm 0.017}$ & ${2.906}^{\pm 0.012}$ & ${13.125}^{\pm 0.121}$ & $2.238^{\pm 0.049}$  \\
    contact $\rightarrow$ motion & \et{0.514}{0.003} & \et{0.760}{0.003} & \et{0.876}{0.002} & \et{0.711}{0.015} & \et{2.863}{0.011} & \et{13.306}{0.133} & \et{2.204}{0.054}\\
    \bottomrule
    \end{tabular}
    }
\end{table*}

\begin{figure*}[tb]
    \centering
    \includegraphics[width=0.8\linewidth]{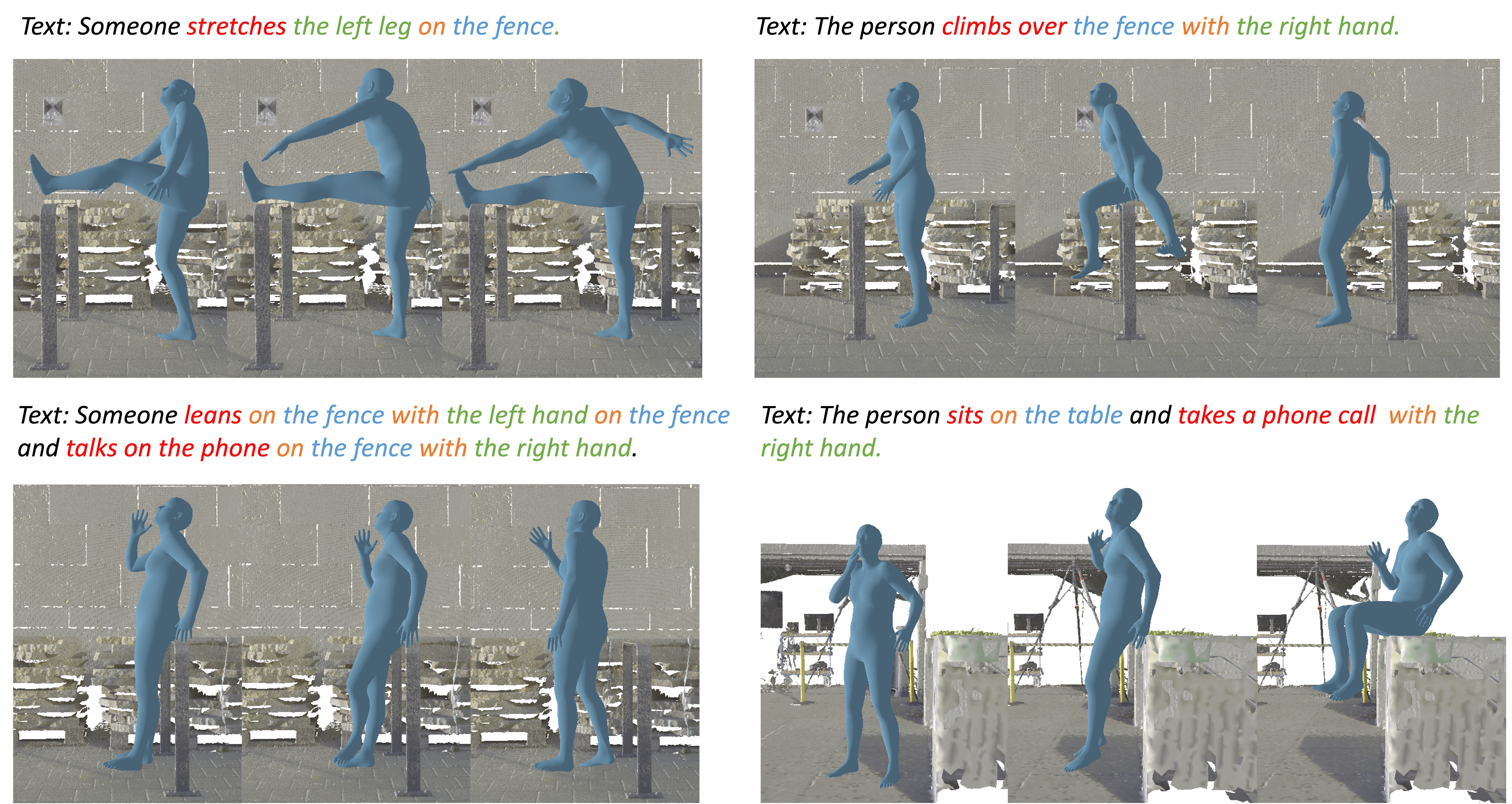}
    \caption{Qualitative results of our method adapted to HOI synthesis. The visual results validate that our method's versatility in adapting to HOI generation.}
    \label{fig:application}
\end{figure*}

\section{Application to HOI Synthesis}
While our method focuses on the generation of interactive human motions conditioned on text, it can be readily extended to the task of Human-Object Interaction (HOI) generation by incorporating an object geometry module. 

Human-object interaction (HOI) generation ~\cite{pmp, interdiff, omomo, cghoi, hoidiff, moveasyousay} aims to produce a holistic human and object motion sequence conditioned on the object's geometry, sparse object motion evidence, and optionally a textual description. 
Our method exhibits versatility in adapting to HOI generation in static scenes. Through the integration of a conditioning module for object geometry, our approach adeptly generates natural and physically plausible motion sequences that interact with the given static objects. 
To handle the object geometry represented as an object mesh, we utilize the Point Transformer~\cite{pointtrans}, pretrained on ScanNet~\cite{scannet}. Initially, the object mesh with N vertices is uniformly sampled down to a fixed length of 1024 points. Subsequently, the Point Transformer produces a 512-dimensional feature vector. This object feature, when concatenated with the text embeddings, serves as input conditions for the intertwined GPT module. This setup enables the module to predict realistic motion and contact tokens that align with both the object and textual inputs.
Visual results are presented in Fig.~\ref{fig:application}. Please refer to the supplementary video demo for dynamic visual results.

\section{Limitation and Future Work}\label{sub:supp_limitation}
Although our method effectively generates realistic and diverse motion sequences based on textual inputs, several limitations warrant attention in future research. Firstly, accurately modeling delicate hand actions poses a challenge. Despite the current model's adept handling of whole-body contacts, finer hand movements sometimes fail to precisely align with object geometries, necessitating separate modeling of hand-action contacts. Secondly, the current framework is tailored for static objects and lacks provisions for dynamic objects that move alongside humans. The integration of these diverse object types within a unified HOI framework presents a significant avenue for further investigation. Lastly, despite the introduction of an efficient automatic text annotation pipeline, minor disparities between annotations and natural language may persist. Delving into the impact of these differences on the model and exploring advanced techniques, such as leveraging large language models, to elevate the quality of automatic annotation, stands as a promising path for exploration.


\section{Conclusion}
In this work, we introduce a novel task of generating 3D interactive human motion from text. We construct a new dataset named \datasetname, derived from RICH dataset~\cite{rich}, featuring high-quality motion data, precise human-object contact labels, and interactive textual descriptions. Leveraging \datasetname, we present \modelname, a novel approach for text-driven interactive human motion synthesis, emphasizing body contact as key evidence. Our method involves the independent modeling of motion and contact through VQ-VAEs, coupled with conditioned generation via an intertwined GPT. Additionally, an interaction-aware text encoder is employed to enhance the understanding of interactive textual descriptions. Experiments showcase the superior performance of our method in generating accurate and context-aware motion sequences from textual input. We also demonstrate the adaptability of our model for HOI generation.

\ifCLASSOPTIONcaptionsoff
  \newpage
\fi

\bibliographystyle{IEEEtran}
\bibliography{main}

\begin{thebibliography}{10}
\providecommand{\url}[1]{#1}
\csname url@samestyle\endcsname
\providecommand{\newblock}{\relax}
\providecommand{\bibinfo}[2]{#2}
\providecommand{\BIBentrySTDinterwordspacing}{\spaceskip=0pt\relax}
\providecommand{\BIBentryALTinterwordstretchfactor}{4}
\providecommand{\BIBentryALTinterwordspacing}{\spaceskip=\fontdimen2\font plus
\BIBentryALTinterwordstretchfactor\fontdimen3\font minus \fontdimen4\font\relax}
\providecommand{\BIBforeignlanguage}[2]{{%
\expandafter\ifx\csname l@#1\endcsname\relax
\typeout{** WARNING: IEEEtran.bst: No hyphenation pattern has been}%
\typeout{** loaded for the language `#1'. Using the pattern for}%
\typeout{** the default language instead.}%
\else
\language=\csname l@#1\endcsname
\fi
#2}}
\providecommand{\BIBdecl}{\relax}
\BIBdecl

\bibitem{rich}
C.-H.~P. Huang, H.~Yi, M.~H{\"o}schle, M.~Safroshkin, T.~Alexiadis, S.~Polikovsky, D.~Scharstein, and M.~J. Black, ``Capturing and inferring dense full-body human-scene contact,'' in \emph{Proceedings of the IEEE/CVF Conference on Computer Vision and Pattern Recognition}, 2022, pp. 13\,274--13\,285.

\bibitem{humanml3d}
C.~Guo, S.~Zou, X.~Zuo, S.~Wang, W.~Ji, X.~Li, and L.~Cheng, ``Generating diverse and natural 3d human motions from text,'' in \emph{Proceedings of the IEEE/CVF Conference on Computer Vision and Pattern Recognition}, 2022, pp. 5152--5161.

\bibitem{language2pose}
C.~Ahuja and L.-P. Morency, ``Language2pose: Natural language grounded pose forecasting,'' in \emph{2019 International Conference on 3D Vision (3DV)}.\hskip 1em plus 0.5em minus 0.4em\relax IEEE, 2019, pp. 719--728.

\bibitem{motionclip}
G.~Tevet, B.~Gordon, A.~Hertz, A.~H. Bermano, and D.~Cohen-Or, ``Motionclip: Exposing human motion generation to clip space,'' in \emph{European Conference on Computer Vision}.\hskip 1em plus 0.5em minus 0.4em\relax Springer, 2022, pp. 358--374.

\bibitem{motiondiffuse}
M.~Zhang, Z.~Cai, L.~Pan, F.~Hong, X.~Guo, L.~Yang, and Z.~Liu, ``Motiondiffuse: Text-driven human motion generation with diffusion model,'' \emph{arXiv preprint arXiv:2208.15001}, 2022.

\bibitem{t2mgpt}
J.~Zhang, Y.~Zhang, X.~Cun, S.~Huang, Y.~Zhang, H.~Zhao, H.~Lu, and X.~Shen, ``T2m-gpt: Generating human motion from textual descriptions with discrete representations,'' \emph{arXiv preprint arXiv:2301.06052}, 2023.

\bibitem{mdm}
G.~Tevet, S.~Raab, B.~Gordon, Y.~Shafir, D.~Cohen-Or, and A.~H. Bermano, ``Human motion diffusion model,'' \emph{arXiv preprint arXiv:2209.14916}, 2022.

\bibitem{mld}
X.~Chen, B.~Jiang, W.~Liu, Z.~Huang, B.~Fu, T.~Chen, and G.~Yu, ``Executing your commands via motion diffusion in latent space,'' in \emph{Proceedings of the IEEE/CVF Conference on Computer Vision and Pattern Recognition}, 2023, pp. 18\,000--18\,010.

\bibitem{imos}
A.~Ghosh, R.~Dabral, V.~Golyanik, C.~Theobalt, and P.~Slusallek, ``Imos: Intent-driven full-body motion synthesis for human-object interactions,'' in \emph{Computer Graphics Forum}, vol.~42, no.~2.\hskip 1em plus 0.5em minus 0.4em\relax Wiley Online Library, 2023, pp. 1--12.

\bibitem{humanise}
Z.~Wang, Y.~Chen, T.~Liu, Y.~Zhu, W.~Liang, and S.~Huang, ``Humanise: Language-conditioned human motion generation in 3d scenes,'' \emph{Advances in Neural Information Processing Systems}, vol.~35, pp. 14\,959--14\,971, 2022.

\bibitem{ghosh2021synthesis}
A.~Ghosh, N.~Cheema, C.~Oguz, C.~Theobalt, and P.~Slusallek, ``Synthesis of compositional animations from textual descriptions,'' in \emph{Proceedings of the IEEE/CVF international conference on computer vision}, 2021, pp. 1396--1406.

\bibitem{temos}
M.~Petrovich, M.~J. Black, and G.~Varol, ``Temos: Generating diverse human motions from textual descriptions,'' in \emph{European Conference on Computer Vision}.\hskip 1em plus 0.5em minus 0.4em\relax Springer, 2022, pp. 480--497.

\bibitem{motiongpt}
B.~Jiang, X.~Chen, W.~Liu, J.~Yu, G.~Yu, and T.~Chen, ``Motiongpt: Human motion as a foreign language,'' \emph{arXiv preprint arXiv:2306.14795}, 2023.

\bibitem{humantomato}
S.~Lu, L.-H. Chen, A.~Zeng, J.~Lin, R.~Zhang, L.~Zhang, and H.-Y. Shum, ``Humantomato: Text-aligned whole-body motion generation,'' \emph{arXiv preprint arXiv:2310.12978}, 2023.

\bibitem{ude}
Z.~Zhou and B.~Wang, ``Ude: A unified driving engine for human motion generation,'' in \emph{Proceedings of the IEEE/CVF Conference on Computer Vision and Pattern Recognition}, 2023, pp. 5632--5641.

\bibitem{mofusion}
R.~Dabral, M.~H. Mughal, V.~Golyanik, and C.~Theobalt, ``Mofusion: A framework for denoising-diffusion-based motion synthesis,'' in \emph{Proceedings of the IEEE/CVF Conference on Computer Vision and Pattern Recognition}, 2023, pp. 9760--9770.

\bibitem{flame}
J.~Kim, J.~Kim, and S.~Choi, ``Flame: Free-form language-based motion synthesis \& editing,'' in \emph{Proceedings of the AAAI Conference on Artificial Intelligence}, vol.~37, no.~7, 2023, pp. 8255--8263.

\bibitem{diffusionmodel}
J.~Sohl-Dickstein, E.~Weiss, N.~Maheswaranathan, and S.~Ganguli, ``Deep unsupervised learning using nonequilibrium thermodynamics,'' in \emph{International conference on machine learning}.\hskip 1em plus 0.5em minus 0.4em\relax PMLR, 2015, pp. 2256--2265.

\bibitem{stablediffusion}
R.~Rombach, A.~Blattmann, D.~Lorenz, P.~Esser, and B.~Ommer, ``High-resolution image synthesis with latent diffusion models,'' in \emph{Proceedings of the IEEE/CVF conference on computer vision and pattern recognition}, 2022, pp. 10\,684--10\,695.

\bibitem{imagen}
C.~Saharia, W.~Chan, S.~Saxena, L.~Li, J.~Whang, E.~L. Denton, K.~Ghasemipour, R.~Gontijo~Lopes, B.~Karagol~Ayan, T.~Salimans \emph{et~al.}, ``Photorealistic text-to-image diffusion models with deep language understanding,'' \emph{Advances in Neural Information Processing Systems}, vol.~35, pp. 36\,479--36\,494, 2022.

\bibitem{dalle2}
A.~Ramesh, P.~Dhariwal, A.~Nichol, C.~Chu, and M.~Chen, ``Hierarchical text-conditional image generation with clip latents,'' \emph{arXiv preprint arXiv:2204.06125}, vol.~1, no.~2, p.~3, 2022.

\bibitem{kitml}
M.~Plappert, C.~Mandery, and T.~Asfour, ``The kit motion-language dataset,'' \emph{Big data}, vol.~4, no.~4, pp. 236--252, 2016.

\bibitem{babel}
A.~R. Punnakkal, A.~Chandrasekaran, N.~Athanasiou, A.~Quiros-Ramirez, and M.~J. Black, ``Babel: Bodies, action and behavior with english labels,'' in \emph{Proceedings of the IEEE/CVF Conference on Computer Vision and Pattern Recognition}, 2021, pp. 722--731.

\bibitem{humanact12}
C.~Guo, X.~Zuo, S.~Wang, S.~Zou, Q.~Sun, A.~Deng, M.~Gong, and L.~Cheng, ``Action2motion: Conditioned generation of 3d human motions,'' in \emph{Proceedings of the 28th ACM International Conference on Multimedia}, 2020, pp. 2021--2029.

\bibitem{uestc}
Y.~Ji, F.~Xu, Y.~Yang, F.~Shen, H.~T. Shen, and W.-S. Zheng, ``A large-scale rgb-d database for arbitrary-view human action recognition,'' in \emph{Proceedings of the 26th ACM international Conference on Multimedia}, 2018, pp. 1510--1518.

\bibitem{nturgbd}
J.~Liu, A.~Shahroudy, M.~Perez, G.~Wang, L.-Y. Duan, and A.~C. Kot, ``Ntu rgb+ d 120: A large-scale benchmark for 3d human activity understanding,'' \emph{IEEE transactions on pattern analysis and machine intelligence}, vol.~42, no.~10, pp. 2684--2701, 2019.

\bibitem{kevin-physics}
K.~Xie, T.~Wang, U.~Iqbal, Y.~Guo, S.~Fidler, and F.~Shkurti, ``Physics-based human motion estimation and synthesis from videos,'' in \emph{Proceedings of the IEEE/CVF International Conference on Computer Vision}, 2021, pp. 11\,532--11\,541.

\bibitem{contact-dynamics}
D.~Rempe, L.~J. Guibas, A.~Hertzmann, B.~Russell, R.~Villegas, and J.~Yang, ``Contact and human dynamics from monocular video,'' in \emph{Computer Vision--ECCV 2020: 16th European Conference, Glasgow, UK, August 23--28, 2020, Proceedings, Part V 16}.\hskip 1em plus 0.5em minus 0.4em\relax Springer, 2020, pp. 71--87.

\bibitem{physcap}
S.~Shimada, V.~Golyanik, W.~Xu, and C.~Theobalt, ``Physcap: Physically plausible monocular 3d motion capture in real time,'' \emph{ACM Transactions on Graphics (ToG)}, vol.~39, no.~6, pp. 1--16, 2020.

\bibitem{dynamics-regulated}
Z.~Luo, R.~Hachiuma, Y.~Yuan, and K.~Kitani, ``Dynamics-regulated kinematic policy for egocentric pose estimation,'' \emph{Advances in Neural Information Processing Systems}, vol.~34, pp. 25\,019--25\,032, 2021.

\bibitem{simpoe}
Y.~Yuan, S.-E. Wei, T.~Simon, K.~Kitani, and J.~Saragih, ``Simpoe: Simulated character control for 3d human pose estimation,'' in \emph{Proceedings of the IEEE/CVF conference on computer vision and pattern recognition}, 2021, pp. 7159--7169.

\bibitem{diffphy}
E.~G{\"a}rtner, M.~Andriluka, E.~Coumans, and C.~Sminchisescu, ``Differentiable dynamics for articulated 3d human motion reconstruction,'' in \emph{Proceedings of the IEEE/CVF Conference on Computer Vision and Pattern Recognition}, 2022, pp. 13\,190--13\,200.

\bibitem{posetriplet}
K.~Gong, B.~Li, J.~Zhang, T.~Wang, J.~Huang, M.~B. Mi, J.~Feng, and X.~Wang, ``Posetriplet: co-evolving 3d human pose estimation, imitation, and hallucination under self-supervision,'' in \emph{Proceedings of the IEEE/CVF Conference on Computer Vision and Pattern Recognition}, 2022, pp. 11\,017--11\,027.

\bibitem{neuralmocon}
B.~Huang, L.~Pan, Y.~Yang, J.~Ju, and Y.~Wang, ``Neural mocon: Neural motion control for physically plausible human motion capture,'' in \emph{Proceedings of the IEEE/CVF Conference on Computer Vision and Pattern Recognition}, 2022, pp. 6417--6426.

\bibitem{gravicap}
R.~Dabral, S.~Shimada, A.~Jain, C.~Theobalt, and V.~Golyanik, ``Gravity-aware monocular 3d human-object reconstruction,'' in \emph{Proceedings of the IEEE/CVF International Conference on Computer Vision}, 2021, pp. 12\,365--12\,374.

\bibitem{shimada-neural}
S.~Shimada, V.~Golyanik, W.~Xu, P.~P{\'e}rez, and C.~Theobalt, ``Neural monocular 3d human motion capture with physical awareness,'' \emph{ACM Transactions on Graphics (ToG)}, vol.~40, no.~4, pp. 1--15, 2021.

\bibitem{lemo}
S.~Zhang, Y.~Zhang, F.~Bogo, M.~Pollefeys, and S.~Tang, ``Learning motion priors for 4d human body capture in 3d scenes,'' in \emph{Proceedings of the IEEE/CVF International Conference on Computer Vision}, 2021, pp. 11\,343--11\,353.

\bibitem{humor}
D.~Rempe, T.~Birdal, A.~Hertzmann, J.~Yang, S.~Sridhar, and L.~J. Guibas, ``Humor: 3d human motion model for robust pose estimation,'' in \emph{Proceedings of the IEEE/CVF international conference on computer vision}, 2021, pp. 11\,488--11\,499.

\bibitem{ma2023grammar}
S.~Ma, Q.~Cao, H.~Yi, J.~Zhang, and D.~Tao, ``Grammar: Ground-aware motion model for 3d human motion reconstruction,'' in \emph{Proceedings of the 31st ACM International Conference on Multimedia}, 2023, pp. 2817--2828.

\bibitem{deco}
S.~Tripathi, A.~Chatterjee, J.-C. Passy, H.~Yi, D.~Tzionas, and M.~J. Black, ``Deco: Dense estimation of 3d human-scene contact in the wild,'' in \emph{Proceedings of the IEEE/CVF International Conference on Computer Vision}, 2023, pp. 8001--8013.

\bibitem{pmp}
J.~Bae, J.~Won, D.~Lim, C.-H. Min, and Y.~M. Kim, ``Pmp: Learning to physically interact with environments using part-wise motion priors,'' \emph{arXiv preprint arXiv:2305.03249}, 2023.

\bibitem{interdiff}
S.~Xu, Z.~Li, Y.-X. Wang, and L.-Y. Gui, ``Interdiff: Generating 3d human-object interactions with physics-informed diffusion,'' in \emph{Proceedings of the IEEE/CVF International Conference on Computer Vision}, 2023, pp. 14\,928--14\,940.

\bibitem{omomo}
J.~Li, J.~Wu, and C.~K. Liu, ``Object motion guided human motion synthesis,'' \emph{ACM Transactions on Graphics (TOG)}, vol.~42, no.~6, pp. 1--11, 2023.

\bibitem{cghoi}
C.~Diller and A.~Dai, ``Cg-hoi: Contact-guided 3d human-object interaction generation,'' in \emph{Proceedings of the IEEE/CVF Conference on Computer Vision and Pattern Recognition}, 2024, pp. 19\,888--19\,901.

\bibitem{hoidiff}
X.~Peng, Y.~Xie, Z.~Wu, V.~Jampani, D.~Sun, and H.~Jiang, ``Hoi-diff: Text-driven synthesis of 3d human-object interactions using diffusion models,'' \emph{arXiv preprint arXiv:2312.06553}, 2023.

\bibitem{pointtrans}
H.~Zhao, L.~Jiang, J.~Jia, P.~H. Torr, and V.~Koltun, ``Point transformer,'' in \emph{Proceedings of the IEEE/CVF international conference on computer vision}, 2021, pp. 16\,259--16\,268.

\bibitem{mmaction}
M.~Contributors, ``Openmmlab's next generation video understanding toolbox and benchmark,'' \url{https://github.com/open-mmlab/mmaction2}, 2020.

\bibitem{smplx}
G.~Pavlakos, V.~Choutas, N.~Ghorbani, T.~Bolkart, A.~A. Osman, D.~Tzionas, and M.~J. Black, ``Expressive body capture: 3d hands, face, and body from a single image,'' in \emph{Proceedings of the IEEE/CVF conference on computer vision and pattern recognition}, 2019, pp. 10\,975--10\,985.

\bibitem{vqvae}
A.~Van Den~Oord, O.~Vinyals \emph{et~al.}, ``Neural discrete representation learning,'' \emph{Advances in neural information processing systems}, vol.~30, 2017.

\bibitem{clip}
A.~Radford, J.~W. Kim, C.~Hallacy, A.~Ramesh, G.~Goh, S.~Agarwal, G.~Sastry, A.~Askell, P.~Mishkin, J.~Clark \emph{et~al.}, ``Learning transferable visual models from natural language supervision,'' in \emph{International conference on machine learning}.\hskip 1em plus 0.5em minus 0.4em\relax PMLR, 2021, pp. 8748--8763.

\bibitem{sbert}
N.~Reimers and I.~Gurevych, ``Sentence-bert: Sentence embeddings using siamese bert-networks,'' \emph{arXiv preprint arXiv:1908.10084}, 2019.

\bibitem{tmr}
M.~Petrovich, M.~J. Black, and G.~Varol, ``Tmr: Text-to-motion retrieval using contrastive 3d human motion synthesis,'' \emph{arXiv preprint arXiv:2305.00976}, 2023.

\bibitem{ctloss}
R.~Hadsell, S.~Chopra, and Y.~LeCun, ``Dimensionality reduction by learning an invariant mapping,'' in \emph{2006 IEEE computer society conference on computer vision and pattern recognition (CVPR'06)}, vol.~2.\hskip 1em plus 0.5em minus 0.4em\relax IEEE, 2006, pp. 1735--1742.

\bibitem{moveasyousay}
Z.~Wang, Y.~Chen, B.~Jia, P.~Li, J.~Zhang, J.~Zhang, T.~Liu, Y.~Zhu, W.~Liang, and S.~Huang, ``Move as you say interact as you can: Language-guided human motion generation with scene affordance,'' in \emph{Proceedings of the IEEE/CVF Conference on Computer Vision and Pattern Recognition}, 2024, pp. 433--444.

\bibitem{scannet}
A.~Dai, A.~X. Chang, M.~Savva, M.~Halber, T.~Funkhouser, and M.~Nie{\ss}ner, ``Scannet: Richly-annotated 3d reconstructions of indoor scenes,'' in \emph{Proceedings of the IEEE conference on computer vision and pattern recognition}, 2017, pp. 5828--5839.

\end{thebibliography}


\begin{IEEEbiographynophoto}{Sihan Ma} is currently pursuing the Ph.D. degree with the School of Computer Science, the University of Sydney under the supervision of Prof. Dacheng Tao and Dr. Jing Zhang. She earned the M.Phil degree from the University of Sydney in 2020 and the bachelor’s degree from Wuhan University in 2018. Her research interests focus on 3D human-centric visual perception and generation, human-object interaction.
\end{IEEEbiographynophoto}

\begin{IEEEbiographynophoto}{Qiong Cao} is a Research Scientist at JD Explore Academy. Before that, she was a Senior Researcher at Tencent. Prior to joining Tencent, she was a Postdoctoral Researcher with the Visual Geometry Group (VGG), Department of Engineering Science, University of Oxford. She obtained her PhD in Computer Science from the University of Exeter. Her research interests lie in computer vision and machine learning, with a specific focus on human-centric 2D and 3D visual perception as well as multi-modal generation. 
\end{IEEEbiographynophoto}

\begin{IEEEbiographynophoto}{Jing Zhang} (Senior Member, IEEE) is currently a Research Fellow at the School of Computer Science, The University of Sydney. He has published more than 60 papers in prestigious conferences and journals, such as CVPR, ICCV, ECCV, NeurlPS, ICLR, IEEE TPAMI, and IJCV. His research interests include computer vision and deep learning. He is also a Senior Program Committee Member of the AAAI Conference on Artificial Intelligence and the International Joint Conference on Artificial Intelligence. He serves as a reviewer for many prestigious journals and conferences.
\end{IEEEbiographynophoto}

\begin{IEEEbiographynophoto}{Dacheng Tao} (F'15) is currently a Distinguished University Professor in the College of Computing \& Data Science at Nanyang Technological University. He mainly applies statistics and mathematics to artificial intelligence and data science, and his research is detailed in one monograph and over 200 publications in prestigious journals and proceedings at leading conferences, with best paper awards, best student paper awards, and test-of-time awards. His publications have been cited over 112K times and he has an h-index 160+ in Google Scholar. He received the 2015 and 2020 Australian Eureka Prize, the 2018 IEEE ICDM Research Contributions Award, and the 2021 IEEE Computer Society McCluskey Technical Achievement Award. He is a Fellow of the Australian Academy of Science, AAAS, ACM and IEEE.
\end{IEEEbiographynophoto}

\vfill
\end{document}